\documentclass[conference]{IEEEtran}
\usepackage[utf8]{inputenc}
\usepackage{hyperref}
\usepackage{comment}
\usepackage{cite}
\usepackage{amsmath,amssymb,amsfonts}
\usepackage{graphicx}
\usepackage{textcomp}
\usepackage{xcolor}
\usepackage[noend]{algpseudocode}
\def\BibTeX{{\rm B\kern-.05em{\sc i\kern-.025em b}\kern-.08em
    T\kern-.1667em\lower.7ex\hbox{E}\kern-.125emX}}
\begin{document}

\title{Genetic Micro-Programs for Automated Software Testing with Large Path Coverage
\thanks{Supported by the NRF Thuthuka Grant Number 13819413.}
}

\author{
 \IEEEauthorblockN{Jarrod Goschen}
 \IEEEauthorblockA{\textit{Department of Computer Science} \\
 \textit{University of Pretoria}\\
 Pretoria, South Africa \\
 u17112631@tuks.co.za}
 \and
 \IEEEauthorblockN{Anna S. Bosman}
 \IEEEauthorblockA{\textit{Department of Computer Science} \\
 \textit{University of Pretoria}\\
 Pretoria, South Africa \\
 anna.bosman@up.ac.za}
 \and
 \IEEEauthorblockN{Stefan Gruner}
 \IEEEauthorblockA{\textit{Department of Computer Science} \\
 \textit{University of Pretoria}\\
 Pretoria, South Africa \\
 sgruner@cs.up.ac.za}
 }

\maketitle

\begin{abstract}

Ongoing progress in computational intelligence (CI) has led to an increased desire to apply CI techniques for the purpose of improving software engineering processes, particularly software testing. Existing state-of-the-art automated software testing techniques focus on utilising search algorithms to discover input values that achieve high execution path coverage. These algorithms are trained on the same code that they intend to test, requiring instrumentation and lengthy search times to test each software component. This paper outlines a novel genetic programming framework, where the evolved solutions are not input values, but micro-programs that can repeatedly generate input values to efficiently explore a software component's input parameter domain. We also argue that our approach can be generalised such as to be applied to many different software systems, and is thus not specific to merely the particular software component on which it was trained.

\end{abstract}


\section{Motivation}
The increase in software prevalence in commercial business as well as further safety-critical application domains (e.g., medical, aviation, etc.) has increased the necessity for software quality assurance. Erroneous program behaviour can indeed lead to severe undesirable consequences \cite{zhivich2009real}. This has put more pressure on software engineers to adequately test their programs. Thereby they spend large portions of their work-time on the analysis of possible run-time execution paths, and on the choice of effective input values by means of which the test path coverage can be maximised. For this purpose control flow graphs are usually generated, on the basis of which the possible paths can be defined \cite{ammannsoftware}, whereby coverage is measured by enumerating the different paths taken during a set of  tests \cite{ahamed2010studying}. 

Discovering \emph{different} run-time paths within such a graph, and identifying representative sets of input values that correspond to those paths, is difficult and time-consuming. Formally: given a software system $S$ with its corresponding graph model $G$ and its input domain $I$, two \emph{different} input values ($i\neq i'$) belong to the \emph{same} sub-domain ($I_j \subset I$) if their run-time paths under test are the \emph{same} ($p_i = p_{i'}$) in $G$. For the sake of work-efficiency in the software engineering processes, such duplication ought to be avoided --- in other words: a newly chosen test input value $i'$ should (ideally) also lead to a hitherto un-visited run-time path $p(i')$ in $G$. 


As far as the automated generation of test input values is concerned, 
a popular approach is \emph{fuzzing} \cite{li2018fuzzing,acostanovel}.
A fuzzer repeatedly generates inputs for the program under test to consume, whereby it is hoped that some faults will be detected by means of unexpected or unusual input values. However, this focus on mere crash discovery is not helpful for the \emph{systematic} identification of all the reasonable test cases (input values) for the program under test. Moreover, fuzzing is also rather inefficient: because the test values are randomly generated, the fuzzer might possibly spend too much time within the same sub-domain $I_j \subset I$ with identical run-time paths $p(i)=p(i')$ in $G$ for all $i,i'$ in $I_j$.

The utilisation of computational intelligence (CI) for the purposes of automated software testing has been investigated with a variety of optimisation techniques\footnote{Indeed, the combination of CI and software engineering (SE) in both directions (SE techniques for CI as well as CI techniques for SE) is recently gaining considerable attention, whereby several conference proceedings and special issues of journals have already been dedicated to this hybrid theme, such as the \emph{Software Testing in the Machine Learning Era} Special Issue of the Empirical Software Engineering (EMSE) journal, 2022.}, 
but the research results of which we are currently aware are predominantly focused on the `ad-hoc' generation of input data in a \emph{non-reusable} manner. This diminishes the practicality and industrial adoption of such approaches, as the search techniques must be newly `tailored' for each new program under test. 

Our work aims to evaluate the suitability of genetic programming (GP) as a strategy to support automated software testing in a re-usable manner by allowing the GP-generated solutions (which are small programs themselves) to explore the input domain of a given software under test dynamically, rather than by generating merely a set of `static' input values. The ability of the GP-generated solutions to produce a wide diversity of test values ultimately leads to better path coverage in $G$. Moreover, because the GP-generated solutions are small programs themselves, they can also be stored and newly applied to new software systems to be tested in future. Our underlying methodological assumption in this approach is, indeed, `inductive': \emph{if} our GP-generated solutions were `good enough' to produce adequate test inputs and coverage percentages for programs under tests with \emph{known} graphs and paths (`white-box' for scientific research purposes), then they `ought to' produce similarly adequate test inputs and coverage percentages also (later) for programs under test with \emph{unknown} graphs and paths (`black-box' for industrial application purposes). Of course, it is not the aim of this work-in-progress paper to provide the ultimate method to achieve all the above-mentioned goals. Rather, we want to explore in a `pilot study' to what extent the proposed technique efficiently `traverses' the various sub-domains $I_j \subset I$, whether this technique has benefits in comparison against already existing solutions, and whether it can be reliably applied to new (hitherto un-seen) software testing problems without significant re-calibration efforts. 


The remainder of this paper is structured as follows: Section~\ref{sec:bg} provides the necessary background. Section \ref{sec:method} discusses the proposed method in some detail. Section \ref{sec:results} presents our empirical results. Section \ref{sec:summary} concludes with some remarks about possible avenues for future work. 

\section{Related Work}\label{sec:bg} 
This section provides the background relevant to this study. Section~\ref{sub:auto} introduces the field of automated software testing, 
and Section~\ref{sub:genp} gives a brief discussion of GP.
\subsection{Automated software testing}\label{sub:auto}\label{sub:search}
For the automation of software testing, much research is focused on the generation of test data. Model-based, randomised, and search-based testing are three of the best-known approaches. Randomised automated testing techniques have have been used for detecting concurrency faults \cite{koushik07rand}, for testing object-oriented software \cite{Ilinca08ART}, as well as for the above-mentioned ``fuzzing'' software components whereby random-generated input values are hoped to force test-runs into unanticipated paths with undesirable results \cite{choudhary2015automated}. However, this approach is no longer viable in practice when a software system under test has too many possible run-time paths, or when its most `interesting' input sub-domains are rather sparse. 


In search-based automated test generation, an `intelligent' algorithm is used in order to discover representative test cases systematically (rather than to generate test cases merely randomly) \cite{mcminn2011search}. Problematic for these approaches, however, is the \emph{discrete} input space of typical software systems under test, whereas CI search algorithms often require continuous search spaces. Moreover, such search spaces can possibly contain `plateaus' on which solution-finding algorithms can get stuck~\cite{mcminn2004search}. Moreover, the success of search-based techniques depends strongly on their fitness function design, as well as on the function's suitability for traversing the fitness landscape of the software system under test \cite{harman2004testability}. Multi-point search algorithms, by contrast, are less likely to get stuck in local optima because they sample several areas of the search space simultaneously. In our work, genetic programming is used as a multi-point search technique for automatic test generation.

\subsection{Genetic programming}\label{sub:genp}
Genetic programming (GP) is a heuristic search technique
in which programs are evolved through a process simulating natural selection \cite{koza1992genetic}. GP's search mechanism is basically the same as in genetic algorithms (GA), whereby an initial population is generated randomly and then refined over several generations through a selection and modification process. The selection method is responsible for choosing the parents for the next generation; the search progresses by way of removing poorly performing individuals and keeping better performing individuals in the `genetic pool'. The modification process changes the selected individuals such as to further the exploration of the search space.

\begin{figure}[t]
    \centering
    \includegraphics[scale=0.3]{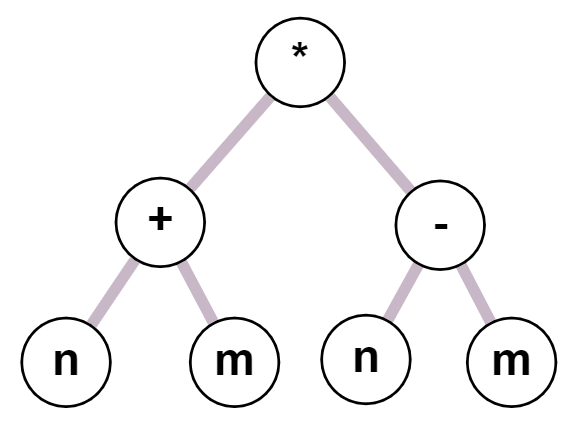}
    \caption{A basic arithmetic program tree, representative of a GP chromosome}
    \label{fig:BasicGP}
\end{figure}

In GP, the evolved individuals (referred to as chromosomes) have a tree structure where each node falls into either the terminal or function set. Terminal nodes act as values and signal the end of a branch, while function nodes operate on one or more successor nodes below it and propagate the result upwards to the ancestor nodes. Fig. \ref{fig:BasicGP} shows a small example of such a program wherein the terminal nodes are placeholders: they can be substituted by actual values before the program (tree) is executed. 

Both GA and GP have been successfully applied for automated test generation, whereby GA appears to outperform the above-mentioned random-input method \cite{sthamer1995automatic,michael2001generating}. 
Moreover, a large \emph{path coverage} percentage for some software systems under test was obtained by means of a GA with binary string chromosomes \cite{pachauri2013automated}. 

GP allows for more complex search landscapes to be traversed due to the increased sophistication of program trees in comparison against simple binary strings. The usage of GP in the domain of automatic software testing is still less explored than the usage of its older GA counterpart. Nevertheless, \cite{huo2017genetic,weimer09GP} point to some advantages of GP (versus GA) in this domain.

\begin{figure}[t]
    \centering
    \includegraphics[scale=0.4]{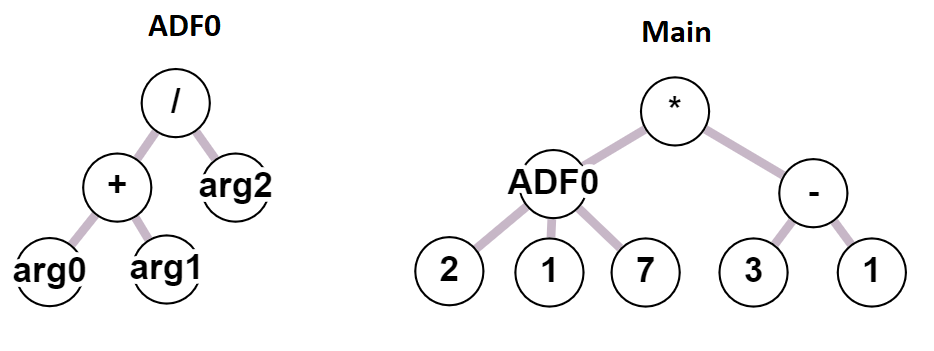}
    \caption{ADF genetic program tree, with one defined function, used in the main program tree}
    \label{fig:ADF}
\end{figure}

Automatically defined functions (ADF) are an extension of 
GP \cite{koza1994genetic2} whereby the generated programs are not merely simple program trees, but complete `micro-programs'. They contain evolved functions which can be used by the result-producing function. Fig. \ref{fig:ADF} shows an example of such a GP individual which uses a single defined function within its result-producing branch. In the context of GP, this approach allows the individuals to generate new functions to increase the already existing functional set. GP implementations with ADF structures have a beneficial `lens effect' \cite{koza1994genetic2} which facilitates the detection of solution-individuals with extreme (very high or very low) scores.

\section{Method}\label{sec:method}

\subsection{Framework}\label{subsec:framework}
\begin{figure}[tb!]
    \centering
    \includegraphics[scale=0.5]{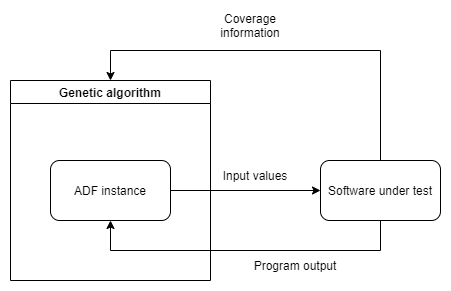}
    \caption{Conceptual framework of our approach}
    \label{fig:Framework}
\end{figure}

Our framework consists of four main elements: a test handler, a GA, genetic micro-programs, and the software under test; (see Fig.~\ref{fig:Framework}). They are explained in the following paragraphs. All code developed for this study is publicly available in the following repository: \url{https://github.com/wolwe1/GP-Automated-Testing/}. 

\subsubsection{Test handler}\label{sub:test_handler}
The first component is the test handler. It `drives' the other components and outputs the experimental results to the user. Thereby the handler must of course comply with the \emph{signature} of the interface of the software under test. This signature information is used to create a generator that can construct the program trees that output the correct number of inputs; also a terminal node is added to the terminal set for storing the last program response. 

\subsubsection{Genetic algorithm}\label{sub:ga}
The next element is the GA for searching the problem space, which is populated by the above-mentioned genetic micro-programs (GMP). The GA initialises the population with randomly generated program trees and uses path coverage information to calculate the individuals' fitness.\footnote{The coverage information must be obtained with help of auxiliary `instrumentation code' placed on relevant paths of the `white-box' software under test.} Our GA uses \emph{tournament selection} which has the desirable property of being \emph{non-elitist} (individuals chosen at random). This strategy permits also somewhat poorly performing individuals to become parents for the next generation if they enter a tournament against individuals whose performance is even worse. This `lenience' is important for maintaining population diversity and for preventing premature convergence of the search in sub-optimal regions. 

As soon as the parents are selected from the population, they are modified by means of several operators. Our implementation uses reproduction, mutation, and crossover. Reproduction maintains the population's stability and approaches solution-convergence by simply passing the individuals to the next generation without further modifications. Mutation drives the search space exploration by selecting a point on the program tree and replacing it with a newly generated sub-tree, where specified mutation depth determines how far down the tree the point of mutation can be chosen. The crossover operator randomly selects two individuals ($x,y$) as well as a sub-tree of each of them ($t_x,t_y$), and then swaps those sub-trees between those individuals (such that $t_x$ is now a sub-tree of $y$ and vice versa). This operation does not merely drive the exploration of the search space, but also enforces information exchange between the individuals involved. Our GA's hyper-parameter values are shown in Table \ref{tab:GAParams}. 

\begin{table}[tb!]
\caption{Parameters for our genetic algorithm}
    \label{tab:GAParams}
\begin{tabular}{ |p{4cm}||p{3.9cm}|  }
  \hline
 Parameter & Value\\
 \hline
 Population size & 150\\
 Number of generations & 100\\
 Number of runs per test case & 5\\
 Reproductive rate & 30\% \\
 Mutation rate & 40\% \\
 Crossover rate & 30\% \\
 Mutation depth & 5 \\
 \hline
\end{tabular}
\end{table}

\subsubsection{Genetic micro-programs}\label{sub:gp}
The individuals of our GA' population (rather than the algorithm as such) are the main concern of this paper. As mentioned above, these individuals are ADFs which generate the input values for the software under test. Each individual is therefore a GMP. The quality of an individual is defined by how much path coverage it can achieve with its generated inputs. Each individual can be executed repeatedly such as to generate a new input set in every repetition. 
In our exploratory pilot-study for this paper, the generator programs were only allowed to produce five input value sets for each test object: larger experiments remain to be carried out in future work. 

\begin{figure}[t]
    \includegraphics[scale=0.2]{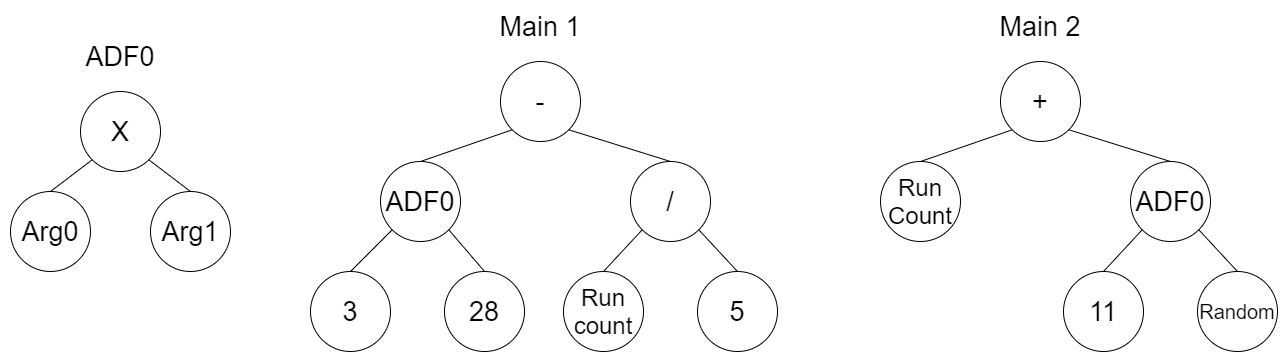}
    \caption{Genetic program that generates two numeric values, with two main programs and a single defined function}
    \label{fig:ADF_example}
\end{figure}

If the interface-signature of the software under test demands multiple inputs (e.g.: $f(i_1, i_2, i_3, \ldots, i_n)$), then the necessary number of GMPs is generated accordingly. In other words: in such cases the generated individuals are not singular programs, but rather \emph{n-tuples} of such programs. Fig. \ref{fig:ADF_example} shows a single GMP individual with two result-producing trees: it is an example of the kind of trees generated for a binary signature with two input parameters (e.g.: $f(i_1, i_2)$). This approach has the advantage of high adaptability, as it can be easily applied to $n$-ary signatures of arbitrarily large $n$ whereby the individual input parameters can even be of different types.


\begin{figure}[t]\centering
    \includegraphics[scale=0.4]{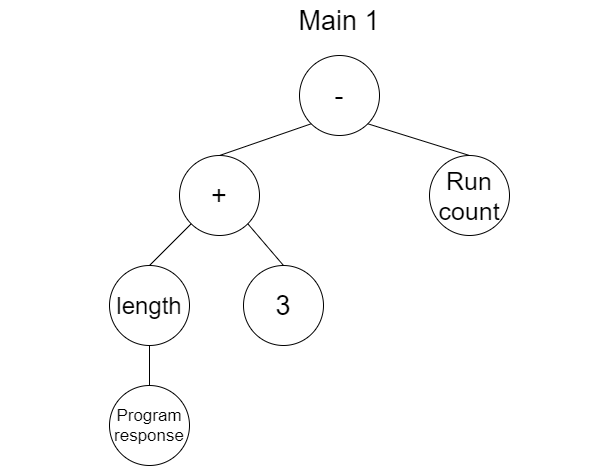}
    \caption{Program tree for a `mixed' signature with strings and number types}
    \label{fig:TypeSwitch}
\end{figure}

The generated programs are correctly typed such as to match with the types in the interface signature of the software under test.
For example, Fig. \ref{fig:TypeSwitch} shows a main program that generates numeric inputs for a software component that returns strings. This example includes a ``length'' function node and propagates according type information through the tree. The use of heterogeneous function nodes thus allows the generated programs to be sufficiently flexible in their construction, which makes them easily applicable to various software components under test.

The construction of the GMPs is controlled by three parameters that influence the complexity of the tree structures. The first is the defined function's maximum depth which corresponds to the permitted height of its tree. Similarly, the \emph{main} program's maximum depth indicates how `high' the result-producing trees of an individual may be. The final parameter is a terminal node's probability of occurrence, which stipulates how likely a chosen node will be a terminal node without any successor nodes.
The higher this probability, the greater the number of branches in the constructed program tree that are pruned in order to meet the maximal depth criterion. For the purposes of our preliminary pilot-study, in which the occurring data structures were not supposed to grow very large, relatively high terminal probabilities were preferred.\footnote{For example: a tree with depth 10 to represent a program consisting entirely of nested \texttt{if-then-else} statements would have more than a  million nodes.} The parameter values used for our GMPs are shown in Table~\ref{tab:GPParams}.\looseness=-1 

\begin{table}[t]
   \caption{Parameters used for genetic micro-program generation}
    \label{tab:GPParams}
    \begin{tabular}{ |p{4cm}||p{4cm}|  }
 \hline
 Parameter & Value\\
 \hline
 Terminal chance & 65\% \\
 Max function depth & 5 \\
 Max main depth & 15 \\
 \hline
 \end{tabular}
\end{table}

Our GMPs are constructed from a predefined set of possible function nodes. Specific functions were chosen to provide sufficient capability for the modification of terminal nodes while still remaining relatively simple in operation. Table \ref{tab:FuncSet} lists the elements in the function node set with an overview of each function. Of note are the \emph{comparator} functions, which act as conditional blocks for use in loops and \texttt{if-then-else} statements. These comparators use entire sub-trees as their predicates, which permits the construction of nested conditional blocks. Table \ref{tab:TermSet} shows the terminal node set. With the exception of the primary value node, all of these nodes are `state-based' in the sense that their values may change from one program execution to another.

\begin{table}[t]
\caption{Functional node set of the genetic micro-programs}
    \label{tab:FuncSet}
    \begin{tabular}{ |p{1.5cm}||p{4.1cm}|p{1.9cm}|  }
 \hline
 Name & description & types supported\\
 \hline
 Add & Addition / concatenation / logical \texttt{and} & string, numeric, Boolean\\
 Division & Safe division & numeric\\
 Multiplication & Multiplication & numeric\\
 Subtraction & Subtraction & numeric\\
 \texttt{if} statement & Executes one of two branches depending on comparator result & string, numeric, Boolean\\
 Loop & Repeats block value until comparator returns false & string, numeric, Boolean\\
 Length of & Returns the length of its child value & string, numeric\\
Equals comparator & Indicates whether its predicates return equal values & string, numeric, Boolean\\
Greater than comparator & Indicates whether its first predicate's value is comparatively larger than the second & numeric, Boolean\\
Less than comparator & Indicates whether its first predicate's value is comparatively smaller than the second & numeric, Boolean\\
Not equal comparator & Indicates whether its first predicate's value differs from the second & string, numeric, Boolean\\
Not null comparator & Indicates whether its predicate's value is null & string\\
\hline
 \end{tabular}
\end{table}

\begin{table}[t]
\caption{Terminal node set of the genetic micro-programs}
    \label{tab:TermSet}
    \begin{tabular}{ |p{1.9cm}||p{3.6cm}|p{2cm}|  }
 \hline
 Name & description & types supported\\
 \hline
 Value & Represents a basic type value: 1-10/'a-z'/true,false & string, numeric, Boolean\\
 Random & Returns a random value from the Value node set & string, numeric, Boolean\\
 Program response & Returns the last response from the {SUT} & string, numeric, Boolean\\
 Output failure & Indicates whether the last output failed & Boolean\\
 Last output & Returns the last value output by the individual & string, numeric, Boolean\\
 Execution count & Returns the number of times the individual has been run & numeric\\
\hline
 \end{tabular}
\end{table}

\subsubsection{Software under test and instrumentation}\label{sub:soft}
The term `software under test' (SUT) refers to the software component being tested. It is assumed that the software has no alterations beyond the injection of instrumentation instructions into its program code such that path coverage information can be obtained. As mentioned above, the path coverage information is essential to guide the search-movements of the GA through the SUT's input domain and to evaluate the fitness of the evolved individuals.

Also note that only the GA uses this path coverage information --- \emph{not} the GMPs in their chromosomes. This feature makes our experimental set-up almost entirely `black-box', as only the SUT's interface and return-values are known to the individuals. Only the initial searching or `learning' phase of the experimental procedure must be `grey-box', because otherwise the GA would not be able to obtain any path coverage information about the SUT (and hence we would not be able to assess the suitability of our method for practical applications in software engineering).

\subsection{Test objects}\label{subsec:test}
For the purposes of our exploratory pilot-study we needed small SUT components with well-understood properties, and therefore did not resort to any publicly available software repositories (benchmarks). Instead, all our small SUT components were implemented ad-hoc (`from scratch'). For example, one of our SUT programs implemented the Euclidian algorithm (with additional path instrumentation), whilst others were designed with more complicated control flow graphs.
All our SUT objects (listed in Table \ref{tab:TestObjects}) have at least been `inspired' by the `characteristics' of similar examples from the literature \cite{windisch2007applying}. In particular, only uniform input type test objects were actually created for the purpose of this paper. However, as mentioned above, mixed input type objects are amenable to our method, too. 

The degree of `difficulty', with which a SUT component challenges our CI analysis method, depends on the SUT's \emph{specificity} as well as on the number of the so-called \emph{prime paths} \cite{ammannsoftware} which its graph model contains\footnote{For the purposes of this paper, we have chosen \emph{prime} path coverage as the criterion for measuring `coverage', though many other coverage criteria are also known \cite{ammannsoftware}. Thereby, a prime path between two nodes is a maximally long simple path which is not a proper sub-path of any other simple path, and a simple path is a path without re-visitation of nodes. Future work could possibly be dedicated to the repetition of our experiments on the basis of other well-known coverage criteria which are defined and explained in the relevant software engineering literature \cite{ammannsoftware}.}. Thereby the specificity of SUT component refers to the number of different input values which lead to the visitation of its prime paths. For example: a SUT component with a conditioned branch, which can only be entered if some $n>10$, has a `lower' specificity than it would have if the same branch's entrance condition would be $20>n>10$. This is because the satisfying input set $\{ i | i>10 \wedge i<20 \}$ is a proper sub-set of $\{ i | i>10 \}$.

\begin{table}[t]
\caption{Description of test objects}
    \label{tab:TestObjects}
    \begin{tabular}{ |p{2.8cm}||p{1.2cm}|p{1.2cm}|p{1.8cm}|  }
 \hline
 Test object name & Input type& No. inputs & No. prime paths\\
 \hline
Palindrome - Iterative & string & 1 & 2 \\
Palindrome - Recursive & string & 1 & 2 \\
Fibonacci - Iterative & numeric & 1 & 2 \\
Fibonacci - Recursive & numeric & 1 & 2 \\
Euclidean - Iterative & numeric & 2 & 2 \\
Euclidean - Recursive & numeric & 2 & 2 \\
Mandelbrot & numeric & 2 & 2 \\
True & Boolean & 1 & 2 \\
True or false & Boolean & 1 & 2 \\
And & Boolean & 2 & 2 \\
Or & Boolean & 2 & 2 \\
And/Or & Boolean & 2 & 3 \\
Xor & Boolean & 2 & 3 \\
Substring & string & 2 & 3\\
Binomial Coefficient & numeric & 2 & 3 \\
Anagram - Recursive & string & 2 & 4 \\
Vowels & string & 2 & 4 \\
Remainder & numeric & 2 & 5 \\
IsPrime & numeric & 1 & 5 \\
Anagram - Iterative & string & 2 & 5 \\
 \hline
 \end{tabular}
\end{table}

\subsection{{GA} training}\label{subsec:train}
For each SUT object, our framework is initialised with the necessary information about the SUT object's signature. Once initialised, the GA generates a population of GMPs and launches the search procedure. In each new generation, each individual is evaluated by means of the fitness function, which repeatedly asks the individual to generate input values for the SUT. Each time the individual is activated, the fitness function invokes the SUT with the generated inputs, collects the SUT's response with the coverage information, and updates any state-based tree nodes in the GMP (such as \emph{last output}, refer to Table~\ref{tab:TermSet}). After the attempt limit is reached, the function takes the cumulative coverage results and identifies the total coverage achieved by the GMP. This information determines the individual's fitness for procreation into the next generation.

In mathematical terms, the fitness of an individual is the percentage of path coverage it achieves on the SUT object across the input set it generated:
$$
C = U\Big(\sum_{i=0}^{n}{
F(x_i)
}\Big)
$$
where $n$ is the number of input sets generated, $x_i$ is a generated input set, $F$ is a coverage result from calling the SUT with $x_i$, and $U$ is a function that calculates which unique nodes have been reached.

The random-generative nature of the GMPs leads to a considerable likelihood of endless or computationally expensive loops in the generated micro-programs. To counteract this problem, the execution of all loops in the generated micro-programs was stopped after 250 iterations, whereby the limit of 250 had been chosen somewhat arbitrarily so that the duration of our run-time experiments remained `manageable'. Any GMP run which reached this limit was regarded as `inconclusive' and no definite coverage percentage information was associated with such a run.

Moreover, just as the \emph{total} input domains vary from one SUT component to the other, so do their computationally \emph{reasonable} sub-domains. For example: Whereas the input of 7 does not cause any trouble with Euclid's algorithm as SUT, the same input 7 would not be the wisest possible choice if the SUT would implement the Ackermann function. Hence, the user of our technique must still have some knowledge about a SUT's \emph{intended} operational \emph{semantics} (not merely its signature) before `blindly' applying our new tools to it.

At the end of the training, the ten best-performing programs for each run with each SUT object were reconstructed. Then we re-created a new population of fifty individuals for each training case. These programs were run with SUT objects that had the same interface (signature) requirements (or lower) than that of the SUT object with which they were trained. Thus, programs trained to produce $n$ input values were tried on SUT objects with \emph{up to} $n$ input parameters, constrained by the return types which, of course, had to be `matching' as well. These SUT objects constituted the generalisation sets which are summarised in Table \ref{tab:Generalisation}. Finally, after the training was completed, they were used to assess the re-usability of the generated GMPs for `new' (hitherto `un-seen') SUTs. 

\begin{table*}[t]
    \caption{Description of the generalisation sets}
    \label{tab:Generalisation}
    \begin{tabular}{ |p{0.25\textwidth}||p{0.7\textwidth}|  }
 \hline
 Training object & Test objects\\
 \hline
And & AndOr, Or, TrueOrFalse, True, Xor \\
True or False & True\\
Or & AndOr, And, TrueOrFalse, True, Xor\\
True & True\\
EuclideanAlgorithm - Iterative & Fibonacci - Iterative, Fibonacci - Iterative, EuclideanAlgorithm - Recursive, BinomialCoefficient\\
EuclideanAlgorithm - Recursive & Fibonacci - Recursive, Fibonacci - Iterative, EuclideanAlgorithm - Iterative, BinomialCoefficient\\
Fibonacci - Iterative & Fibonnaci - Recursive\\
Fibonacci - Recursive & Fibonnaci - Iterative\\
Palindrome - Iterative & Palindrome - Recursive\\
Palindrome - Recursive & Palindrome - Iterative\\
BinomialCoefficient & Fibonacci - Iterative, Fibonacci - Recursive, EuclideanAlgorithm - Iterative, EuclideanAlgorithm - Recursive\\
AndOr & Xor, True, TrueOrFalse, And, Or\\
Xor & True, TrueOrFalse, And, Or, AndOr\\
Substring & Vowels, Anagram - Iterative, Anagram - Recursive\\
Anagram - Iterative & Vowels, Substring, Anagram - Recursive\\
Anagram - Recursive & Vowels, Substring, Anagram - Iterative\\
Remainder & Fibonacci - Iterative, Fibonacci - Recursive, EuclideanAlgorithm - Iterative, EuclideanAlgorithm - Recursive, BinomialCoefficient\\
Vowels & Substring, Anagram - Iterative, Anagram - Recursive\\
 \hline
 \end{tabular}
\end{table*}
\section{Experimental Results}\label{sec:results}

\subsection{Training results}
\begin{figure}[t]
    \includegraphics[width=0.45\textwidth]{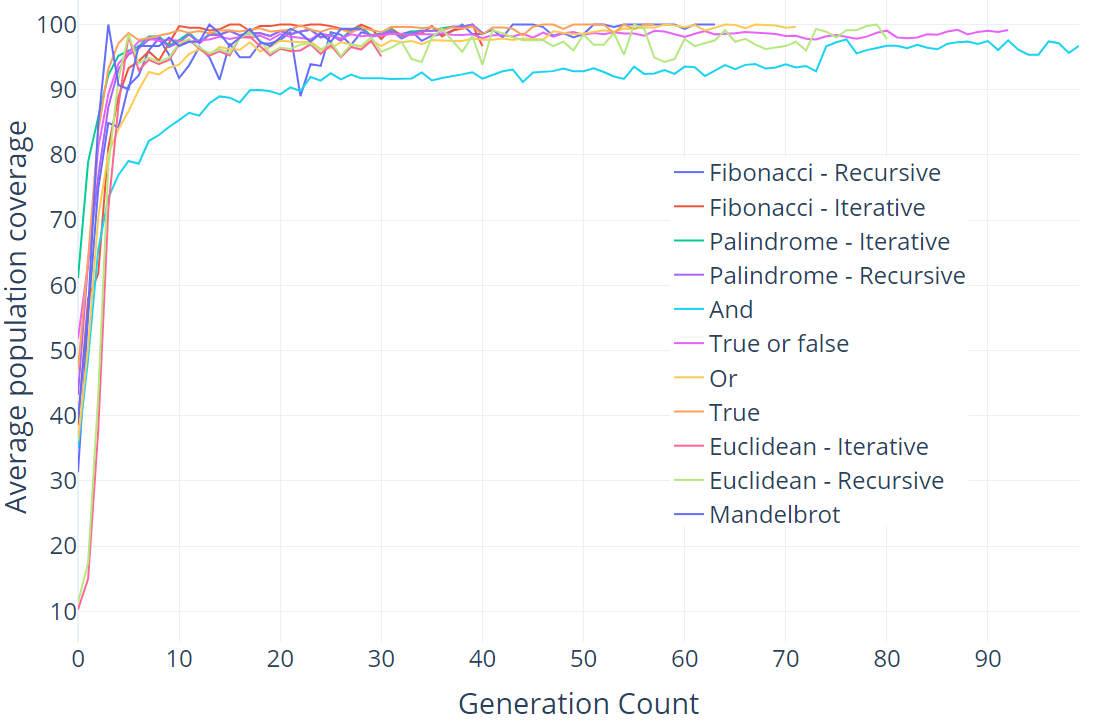}
    \caption{Average population fitness over all runs for individuals trained on 2 prime path containing software components}
    \label{fig:PP2-Train-avg}
\end{figure}

\begin{figure}[t]
    \includegraphics[width=0.45\textwidth]{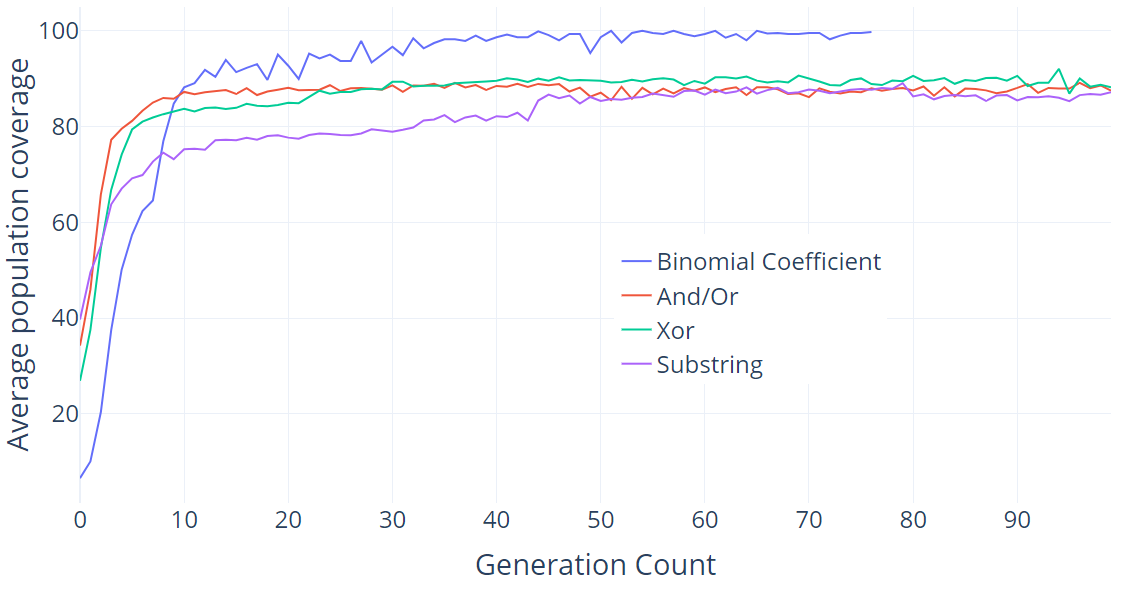}
    \caption{Average population fitness over all runs for individuals trained on {SUT} components with 3 prime paths}
    \label{fig:PP3-Train-avg}
\end{figure}

\begin{figure}[t]
    \includegraphics[width=0.45\textwidth]{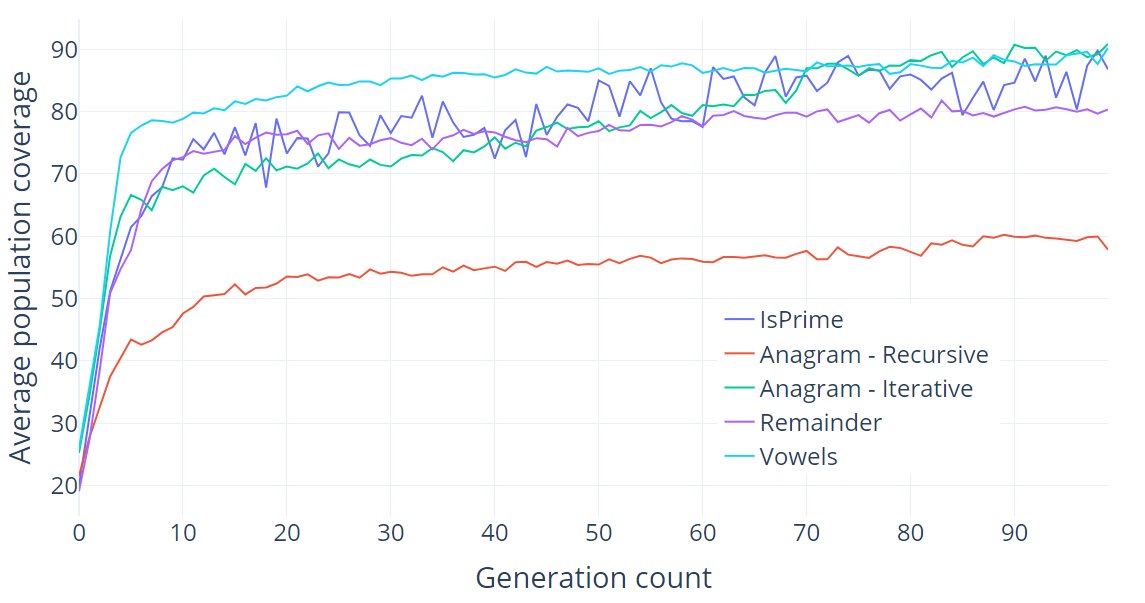}
    \caption{Average population fitness over all runs for individuals trained on {SUT} components with 4--5 prime paths}
    \label{fig:PP45-Train-avg}
\end{figure}

The average population fitness over all runs for GMPs trained on SUTs with 2, 3, and 4--5 number of paths are presented in Fig. \ref{fig:PP2-Train-avg}--\ref{fig:PP45-Train-avg}. Fig. \ref{fig:PP2-Train-avg} shows that for test objects with one or two prime paths, the populations achieved an average of $\approx 85\%$ coverage within 5 generations. We also observed that some individuals achieved 100\% prime path coverage within the first generation.

GMPs trained on test objects with 3 prime paths had marginally slower convergence, but also consistently achieved high coverage: see Fig. \ref{fig:PP3-Train-avg}. On average, the generated micro-programs achieved $\approx 80\%$ coverage within ten generations, whereby some of the generated individuals even achieved 100\% coverage.

The SUTs containing 4--5 prime paths appeared to be somewhat more problematic for our framework. Fewer individuals (than previously) attained 100\% coverage, and the overall population's coverage capability dropped from an average of $\approx 90\%$ to $\approx 80\%$ at the end of the search: see Fig. \ref{fig:PP45-Train-avg}. The worst performance ($\approx 60\%$ coverage) was observed in the experiments with the recursive version of the `Anagram' SUT. In general, the SUTs with 5 prime paths marked the end of 100\% coverage capability for the GMPs: no GMPs generated for any of the 5 prime path SUTs achieved 100\% coverage. 

\begin{figure}[t]
    \includegraphics[width=0.45\textwidth]{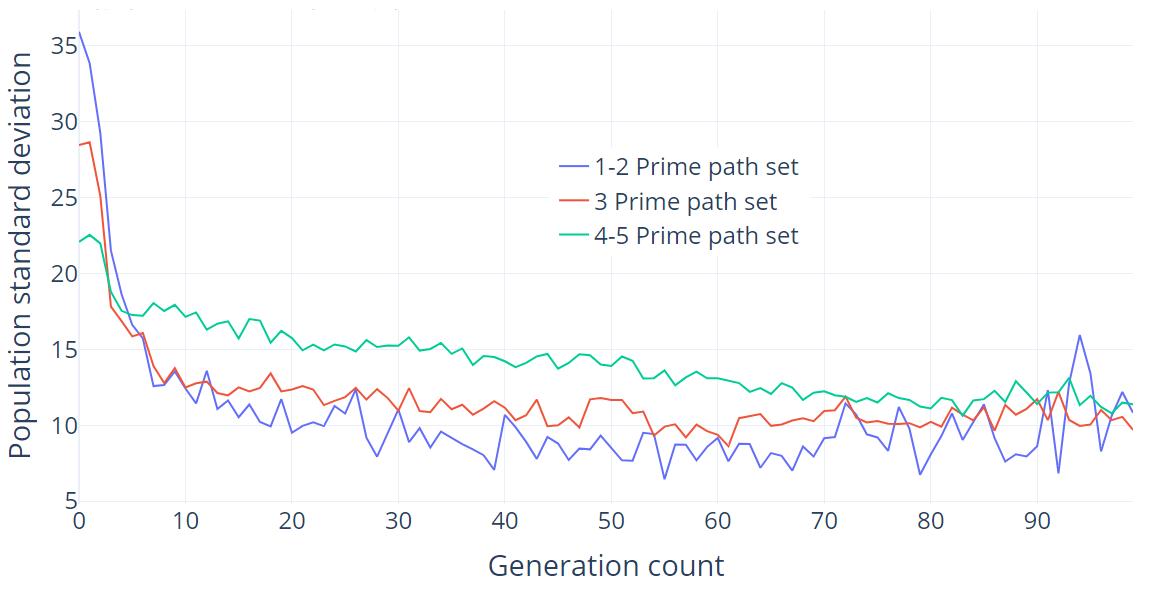}
    \caption{Average of the average population standard deviation for each test case, grouped by SUT complexity}
    \label{fig:StdDev}
\end{figure}

Fig. \ref{fig:StdDev} illustrates the average standard deviation across the generations grouped by SUT complexity. This figure shows that a larger number of prime paths correlated with a higher standard deviation in the early generations; however the test runs converged to values of similar magnitude towards the end of training. 

Analysing the results presented above we can argue that the proposed method showed satisfactory performance. The generated micro-programs achieved high code coverage in most cases considered. The standard deviation results indicate that, although a larger number of prime paths led to slower convergence, the proposed method has not exhibited any divergent behaviour on any of the SUTs considered. 


\subsection{Generalisation results}
After the test cases were completed, the ten best performing GMPs of each run for each SUT were reinitialised. Whereas some of the generated micro-programs were later applied to new SUTs that were structurally quite similar to those SUTs on which those micro-programs were initially trained, we also conducted other experiments in which the generated micro-programs were applied to SUTs that were structurally quite different from the SUTs on which those micro-programs were trained: see Table \ref{tab:Generalisation}. To further investigate the exploratory capability of our technique, the GMPs were also allowed to generate \emph{ten} input sets for the unseen SUTs (in contrast to the training limit of five). Intuitively, one might expect `better' after-training-performance with `similar' after-training-SUTs and `worse' after-training-performance with `different' after-training-SUTs: see below.

\begin{figure*}[t]
\centering
    \includegraphics[width=\textwidth]{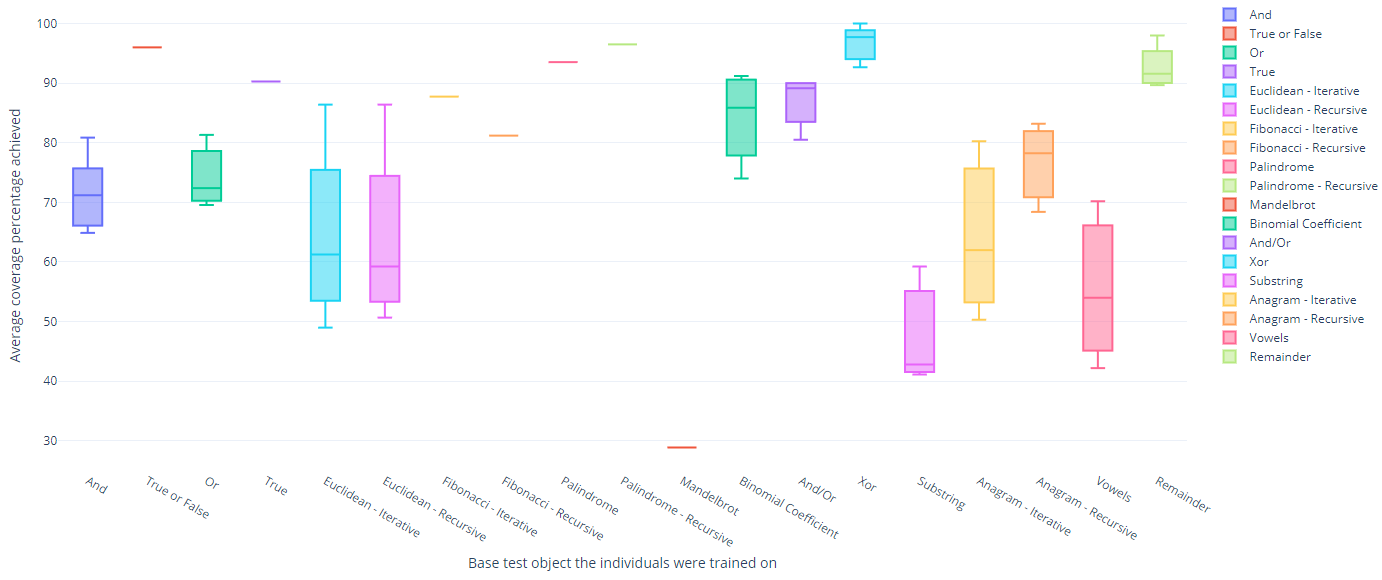}
    \caption{Box plot showing average population fitness against generalisation set, represented by the populations training object}
    \label{fig:boxplot}
\end{figure*}

Fig. \ref{fig:boxplot} shows the generalisation results in the form of a box plot whereby each `box' represents the average and standard deviation of the population's fitness (as described above) when applied to the hitherto unseen SUTs: see Table \ref{tab:Generalisation} for comparison. 

All in all, Fig. \ref{fig:boxplot} shows that the GMPs have indeed \emph{some} ability to generalise (with path coverage $>60\%$ on hitherto unseen SUTs in most cases). 

In line with the `intuitive expectation' mentioned above, Fig. \ref{fig:boxplot} shows indeed a \emph{positive correlation} between the similarity of the generalisation set with the training object on the one hand and the coverage achieved with the training set on the other hand. Populations trained on one version of a function and tested against another version (e.g.: trained on the `palindrome' SUT and tested with its recursive variant, or vice-versa) achieved coverage as high as on their initial training object. Surprisingly, however, the populations trained on more complex SUT objects did \emph{not} exhibit better generalisation capabilities than the ones that had been trained on simpler SUTs. The best generalisation capabilities seem to occur when the generalisation set is similar to the training object, and the training object is more complex. This can be observed with the populations trained on ``Xor'' and ``And/Or'': both contained three prime paths which were tested against the other Boolean functions with simpler path structures. 

Observation of the structures used by successful GMPs did not yield substantial insight. Many contained redundant code or nonsensical constructs. Many successful GMPs used their defined function though others succeeded without them. The program sizes between successful GMPs generated on the same test object also varied by orders of magnitude with little indication of how a program's size related to its capability. 

\section{Conclusion and Outlook to Future Work}\label{sec:summary}
In this paper we proposed a genetic programming (GP) framework with the ultimate goal of improving automated software test generation for black-box systems. Our method is generally suitable for SUT components with arbitrary signatures. Our genetically constructed micro-programs (GMPs) achieved high coverage percentages with a small set of instances that can be stored for later `standalone' re-use.
Their performance, however, somewhat deteriorated for test objects of higher structural complexity; the deeper reasons (and possible remedies) remain to be investigated. The observed variance in path coverage raises questions about how `difficulty' ought to be defined in the context of testing software components, and how this conceptual notion can be made quantitatively measurable in a non-trivial and practically fruitful way.

To date, we cannot indicate how our generated micro-programs would cope with the testing of `realistic' software components that are considerably larger than the ones listed above in Table \ref{tab:TestObjects}. To this end, further and more detailed investigations into the exploratory capability of the generated micro-programs might be useful, whereby the efficacy of the chosen chromosome operators (and other operators) could also be analysed in the hope for insight into the most efficient and effective search procedure. Hereby we would also have to take into account the well-known domain-specific differences of the various `classes' of software systems, instead of hoping to find a `one-size-fits-all' solution for arbitrary kinds of software 
systems.\footnote{For example, the testing of a compiler (software) differs (and must differ) considerably from the testing of a database (software).} 

As it is typical for GP applications, our automatically generated miro-programs contained many useless `introns', such that future work concerning their prevention or removal might be helpful, too. Furthermore, the extension of our framework towards coping with higher-level data-types (e.g.: lists) instead of merely `atomic' data-types (such as numbers or strings) might increase the applicability of our technique to larger classes of SUT systems which consume those composite data-types for input. The applicability of linear genetic programming (LGP) to micro-program generation can also be investigated in future. Finally, a critical comparison of the proposed technique to the state-of-the-art automated software testing solutions needs to be carried out.
\section*{Acknowledgements}
This study was supported by the NRF Thuthuka Grant Number 13819413.
\bibliographystyle{IEEEtran}
\bibliography{refs}
\end{document}